%% file: main.tex
\pdfoutput=1

\documentclass[11pt]{article}

\usepackage{EMNLP2023}

\usepackage{times}
\usepackage{latexsym}

\usepackage[T1]{fontenc}

\usepackage[utf8]{inputenc}

\usepackage{microtype}

\usepackage{inconsolata}

\usepackage[utf8]{inputenc} 
\usepackage[T1]{fontenc}    
\usepackage{hyperref}       
\usepackage{url}            
\usepackage{booktabs}       
\usepackage{amsfonts}       
\usepackage{nicefrac}       
\usepackage{microtype}      
\usepackage{xcolor}         
\usepackage{graphicx}         
\usepackage[english]{babel}
\usepackage{xspace}
\usepackage{multirow}
\usepackage{booktabs}
\usepackage{amsmath}
\usepackage{caption}
\usepackage{subcaption}
\usepackage{capt-of}
\usepackage[]{graphicx}
\usepackage{multirow}
\usepackage{arydshln}
\usepackage{comment}

\usepackage{tabularray}

\newcommand{\approach}[0]{NL2VI\xspace}
\newcommand{\ex}[1]{``\emph{#1}''}

\title{Transferring Visual Attributes from \\Natural Language to Verified Image Generation}

\newcommand*{\affaddr}[1]{#1} 
\newcommand*{\affmark}[1][*]{\textsuperscript{#1}}
\newcommand*{\email}[1]{\texttt{#1}}

\author{Rodrigo Valerio\affmark[1], Joao Bordalo\affmark[1], Michal Yarom\affmark[2], Yonatan Bitton\affmark[2, 3], Idan Szpektor\affmark[2], Joao Magalhaes\affmark[1]\\
\affaddr{\affmark[1]Universidade NOVA de Lisboa}\\
\affaddr{\affmark[2]Google Research}\\
\affaddr{\affmark[3]The Hebrew University of Jerusalem}\\
\email{\{r.valerio, j.bordalo\}@campus.fct.unl.pt}\\
}

\begin{document}

\maketitle
\begin{abstract}
Text to image generation methods (T2I) are widely popular in generating art and other creative artifacts. While visual hallucinations can be a positive factor in scenarios where creativity is appreciated, such artifacts are poorly suited for cases where the generated image needs to be grounded in complex natural language without explicit visual elements.
In this paper, we propose to strengthen the consistency property of T2I methods in the presence of \textit{natural complex language}, which often breaks the limits of T2I methods by including non-visual information, and textual elements that require knowledge for accurate generation (see Figure~\ref{fig:hallucinations_eggs}).
To address these phenomena, we propose a \textbf{N}atural \textbf{L}anguage to \textbf{V}erified \textbf{I}mage generation approach (\textbf{\approach}) that converts a natural prompt into a \textit{visual prompt}, which is more suitable for image generation.
A T2I model then generates an image for the visual prompt, which is then verified with VQA algorithms.
Experimentally, aligning natural prompts with image generation can improve the consistency of the generated images by up to 11\% over the state of the art.
Moreover, improvements can generalize to challenging domains like cooking and DIY tasks, where the correctness of the generated image is crucial to illustrate actions.

\end{abstract}

\section{Introduction}
\begin{figure}[t]
    \centering
    \begin{subfigure}[b]{\linewidth}
        \centering
        \textbf{\small Natural Language Prompt: }\textit{\small "How to purchase company shares: With the ease of online investing, buying shares of a company has become a relatively simple way to build a nest egg or start a retirement fund."}
        \includegraphics[width=\linewidth]{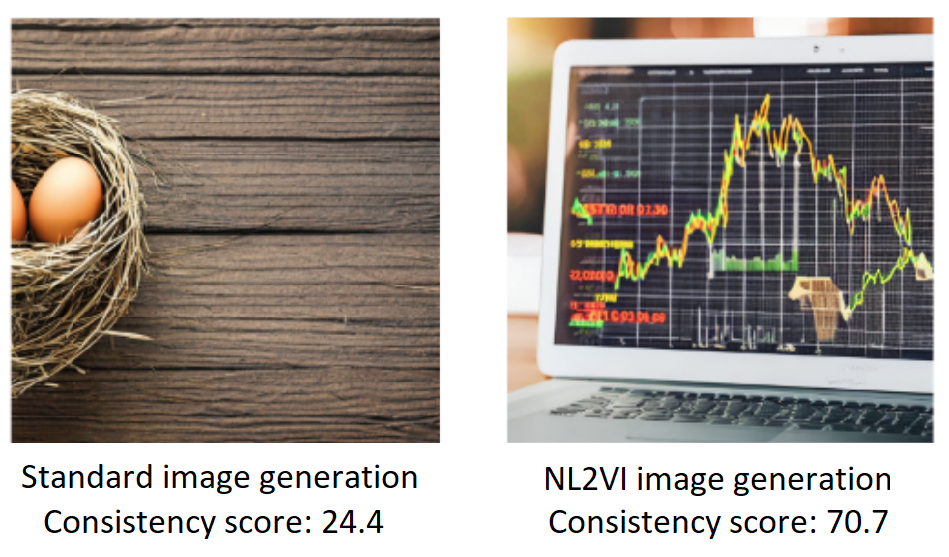}\\
    \end{subfigure}
    \caption{NL2VI can transfer the visual aspects of natural language into generated images and measure its consistency score.}
    \label{fig:hallucinations_eggs}
\end{figure}

Text-to-image generation (T2I) methods~\cite{dall-e2, stable-diffusion, imagen, Muse} are able to map textual prompts to latent image representations in order to represent objects, actions, scenes or emotions mentioned in the prompt. Yet, these models still often produce inconsistencies between the prompt and the image \cite{park2021benchmark, dall-e2.fails.synthatic}, as well as hallucinations that escape visual common sense knowledge, e.g. left side of Figure~\ref{fig:hallucinations_eggs}.
Visual and common sense inconsistencies are further exacerbated when the input is a natural text, instead of direct explicit drawing instructions. This is because natural texts often include information which is not depictable in images, such as emotions and domain knowledge inference e.g., \ex{How to buy company shares}.
This creates a reliability problem when deploying T2I methods in scenarios where correctness is key, e.g. illustration of actions or data augmentation~\cite{data-augmentation-diffusion}.

Previous attempts to verify the consistency of generated images have many limitations. Initial approaches \cite{dalleval, gokhale2022benchmarking} relied on heuristics extracted from image metadata together with an object detector. Recently, \cite{tifa} generates questions with a large language model paired with Visual Question Answering (VQA) for benchmarking T2I models. However, it is important to note that the effectiveness of TIFA~\cite{tifa} is limited to simple and descriptive prompts.

\begin{figure*}[t]
    \centering
    \includegraphics[width=1.0\textwidth]{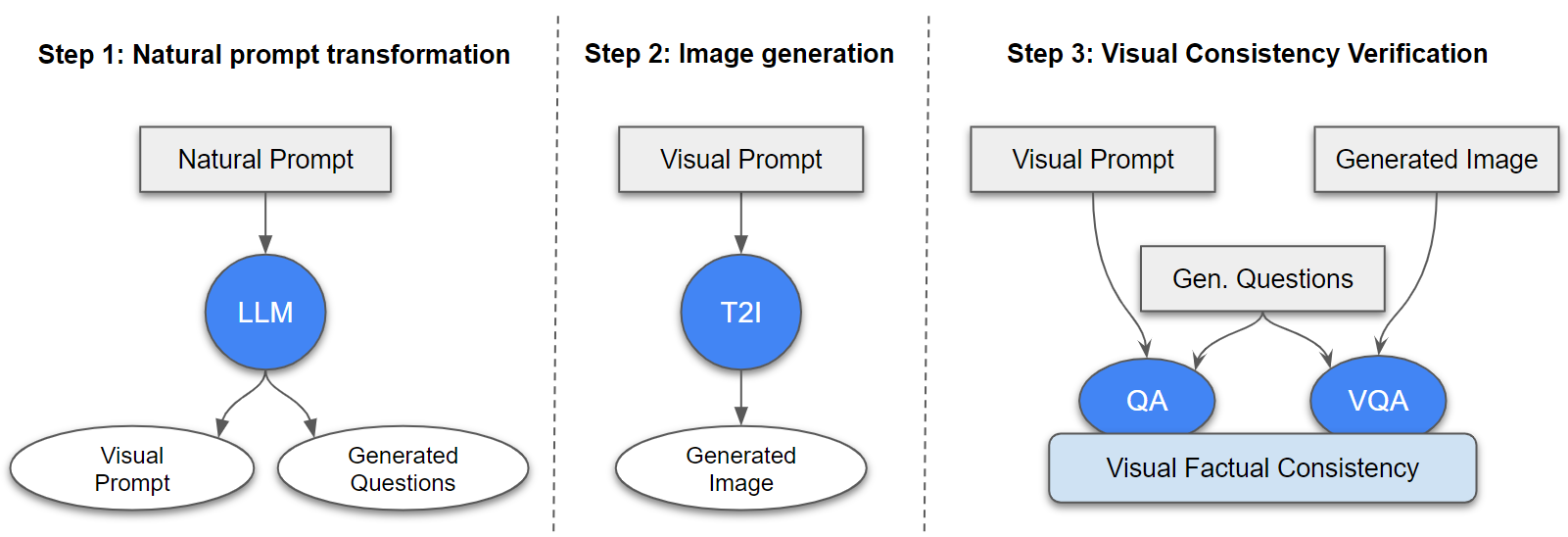}
    \caption{Natural language to verified image generation (\approach).}
    \label{fig:framework}
\end{figure*}

In this paper, we move beyond synthetic and caption-based prompts and aim to \textit{align image generation with natural prompts, while trying to guarantee the visual consistency of the generated image}.
Concretely, we propose a \textbf{N}atural \textbf{L}anguage to \textbf{V}erified \textbf{I}mage generation approach (\approach). \approach first converts natural text into visually plausible text, which we refer to as a \textit{visual prompt}. This conversion is achieved by employing a few-shot LLM that rewrites the natural text, removing non-visual aspects and providing details for textual information that would require common or domain knowledge inference to accurately image generation. Following that, we use a T2I method to generate candidate images for the visual prompt. Finally, the candidate images are ranked using a consistency verification module, and the top ranked image is offered as output.

We show that our method has significant gains on natural prompts over previous methods. Our testing demonstrates that the visual prompt depicts the visual elements in the natural prompt correctly, with an AUC of over 94\%. 
Moreover, our method can predict the consistency of an image with an accuracy 7.8\% and 11.0\% higher on the recipes and DIY domains compared to the state of the art.

\section{Aligning Natural Prompts with Image Generation}
To improve the alignment between \textit{natural prompts} and generated images, we propose the \approach approach to transform the natural prompt into a \textit{visual prompt}, in which all the visual elements present in the natural prompt are well identified and complete, while non-visual elements are removed. We hypothesize that visual prompts would narrow the gap that T2I models need to bridge between the natural text input and the resulting image, and reduce visual hallucinations or generation of implausible artifacts.
 
Concretely, we propose the process depicted in Figure~\ref{fig:framework}. It includes three phases, which rely on recent advances of LLMs and VQA. 
In the first phase, a large language model, e.g. PaLM~\cite{palm} or GPT-3.5, is instructed to distill a visual prompt from an input natural prompt, as well as explicitly indicate the main visual aspects that need to be verified in a generated image as a list of question/answer pairs. 
In the second phase, a conditioned T2I model is asked to follow the generated textual instructions to generate the image.
Finally, in the third phase, a Visual Question Answering (VQA) model provides the answers to the textual questions generated in the first phase based on the generated images.
VQA answers are then compared to the expected textual answers with to detect inconsistencies between the natural prompt and the generated image.

When all these aspects are taken into account, we obtain a method that offers guarantees of generating an image that is aligned with a natural prompt, even when that prompt provides no visual clues. In the following Sections, we detail each of these phases.

\subsection{Visual Prompt and Question Generation for Visual Consistency}
Modern LLMs have been trained on very large corpora and across a wide range of tasks. Leveraging this rich training data, LLMs are able to detect which elements of a textual passage are visually transferable into an image. Figure~\ref{fig:hallucinations_eggs} illustrates how \textit{"purchasing company shares"} is transferred into a \textit{"computer screen with market shares"}.
We build on this property of LLMs to translate a natural prompt into a visual prompt, and explore in-context few-shot learning to generate a visual prompt, see Table \ref{tab:in-context}.
Additionally, we aim to verify that a generated image, conditioned on the generated visual prompt, follows the prompt instruction accurately. To this end, we also instruct the LLM with in-context few-shot examples to generate a set of question/answer pairs that will be used to verify the alignment between the text and the generated image, see Table \ref{tab:in-context}. 
We have two types of questions: binary questions, which serve to verify the presence of objects from the prompt in the image, and open-ended questions, which are more general. The distribution of questions is further detailed in Table~\ref{tab:question-distribution}.

Finally, we note that since this phase is a generative process, it could also be prone to hallucinations and inconsistencies.
To address potential inconsistencies in the generative process, we utilize a commonly accepted method of employing generated question/answer pairs~\cite{q2,nli,true}.
These question-answer pairs are used to validate the consistency of the generated images by evaluating if the VQA answers, based on the visual prompt and the generated images, are in accordance with the QA answers based solely on the visual prompt.
We discuss this in further detail in Sections~\ref{sec:verification} and \ref{sec:ablation_qa_vqa}.

\begin{table}[t]
\centering
\scriptsize
\tiny
\begin{tabular}{|p{0.9\linewidth}|}
\hline
\\
\texttt{As an AI Image Verification Specialist, your primary responsibility is to create a text2img prompt and examine its accuracy assuming an associated image. Your task involves two main steps:}\\
\\
\texttt{Construct a text2img prompt: You will be provided with a description. It's crucial that your text2img prompt incorporates all visual aspects mentioned in the description. The text2img prompt must be a detailed visual description that accurately represents the visual attributes of the description. Non-visual attributes should not be included.}\\
\\
\texttt{Formulate a series of questions: Your questions must be related to the visible components within the image.  Questions must be simple, unambiguous, and answerable based on the observable content in the image. Questions must be about elements that exist in the image and are clearly visible. Questions must be about a single element.}\\
\\
\texttt{Follow the examples below and complete:}\\
\\
\texttt{Description: "Mango-Black Bean Salsa. This fiery, flavorful salsa is excellent with tortilla chips, spooned over roasted meat or fish, or as a topping for quesadillas. Made with mango, avocado, no-salt-added black beans, red onion, jalapeño pepper, cilantro, lime, salt."}\\
\\
\texttt{text2img prompt: A bowl of mango and black bean salsa with tortilla chips. The salsa also has onions, jalapeño peppers and cilantro.}\\
\texttt{\begin{tabular}{l}
Q: is there a bowl of food? A: yes\\
Q: is there salsa? A: yes\\
Q: are there black beans in the salsa?   A: yes\\
Q: Are there mangos in the salsa?        A: yes\\
Q: are there tortilla chips?             A: yes\\
Q: is there cilantro?                    A: yes\\
\end{tabular}}\\
\textit{. . . . . .}\\
\textit{\color{gray}(in context learning examples)}\\
\textit{. . . . . .}\\
\textit{\color{gray}(unseen example)}\\
\texttt{Description: "Garlic Parmesan Pasta. The hardest part is chopping the parsley. Made with: parsley, garlic, butter, chicken broth, milk, parmesan cheese, salt, ground pepper.}\\
\\
\texttt{\color{blue}text2img prompt: A bowl of garlic parmesan pasta with parmesan cheese and parsley.}\\
\texttt{\color{blue}Questions:}\\
\texttt{\color{blue}\begin{tabular}{l}
Q: what is in the bowl?             A: pasta\\
Q: is there a bowl of food?         A: yes\\
Q: is there cheese?                 A: yes\\
Q: is there cheese on the pasta?    A: yes\\
Q: is there parsley?                A: yes\\
\end{tabular}}\\
\\
\hline
\end{tabular}
\caption{\label{tab:in-context} In-context instructions used to transform a natural prompt into a visual prompt and a set of visual consistency evaluation questions.}
\end{table}

\subsection{Text to Image Generation}
In this phase, a T2I model is conditioned on the visual prompt, with more visual details and with fewer ambiguous descriptions, step 2 of Figure~\ref{fig:framework}. 
All visual prompts were shorter than the input limit of the T2I methods we tested, hence no truncation happened.

\subsection{NL2VI Consistency Verification}
\label{sec:verification}
The final phase in \approach is to verify the consistency of the generated image with the visual prompt.
The rationale is to check the consistency of the visual prompt with respect to the natural prompt, then the generated questions and lastly, the generated image.
Inspired by~\citet{true} and~\citet{tifa}, we leverage the questions generated in phase 1 to probe both the image and the prompt and assess the consistency of the answers. To filter the questions based on the prompt, we utilize a Question Answering (QA) model~\cite{unifiedqa} in conjunction with a Natural Language Inference (NLI) model~\cite{nli-model}. 
Although we investigated various methods to verify the consistency between the natural prompt and the visual prompt, the fact is that the performance of the LLM has an AUC of over 90\% which makes it sufficiently robust to consider that they are correct in general.

We are then left with the task of checking the consistency of the generated images with the questions generated by the LLM. For that purpose, we use VQA models, fine-tuned on the VQAv2~\cite{vqa2_dataset} dataset, to answer the questions based on the generated image. Note that, although VQAv2 is a closed domain answer dataset, we use generative models, which may cause open-answers, in particular when there is a clear mismatch between the questions and the input text.
Hence, to compare the answers, we tested several text matching algorithms, including string equality, BERTScore~\cite{bertscore} and NLI~\cite{nli}.

\section{Experimental Methodology}
In this Section, we present the NL2VI implementation details and describe the experimental setup used to evaluate NL2VI.

\subsection{Implementation}
\begin{figure}[t]
    \centering
    \includegraphics[width=0.49\textwidth]{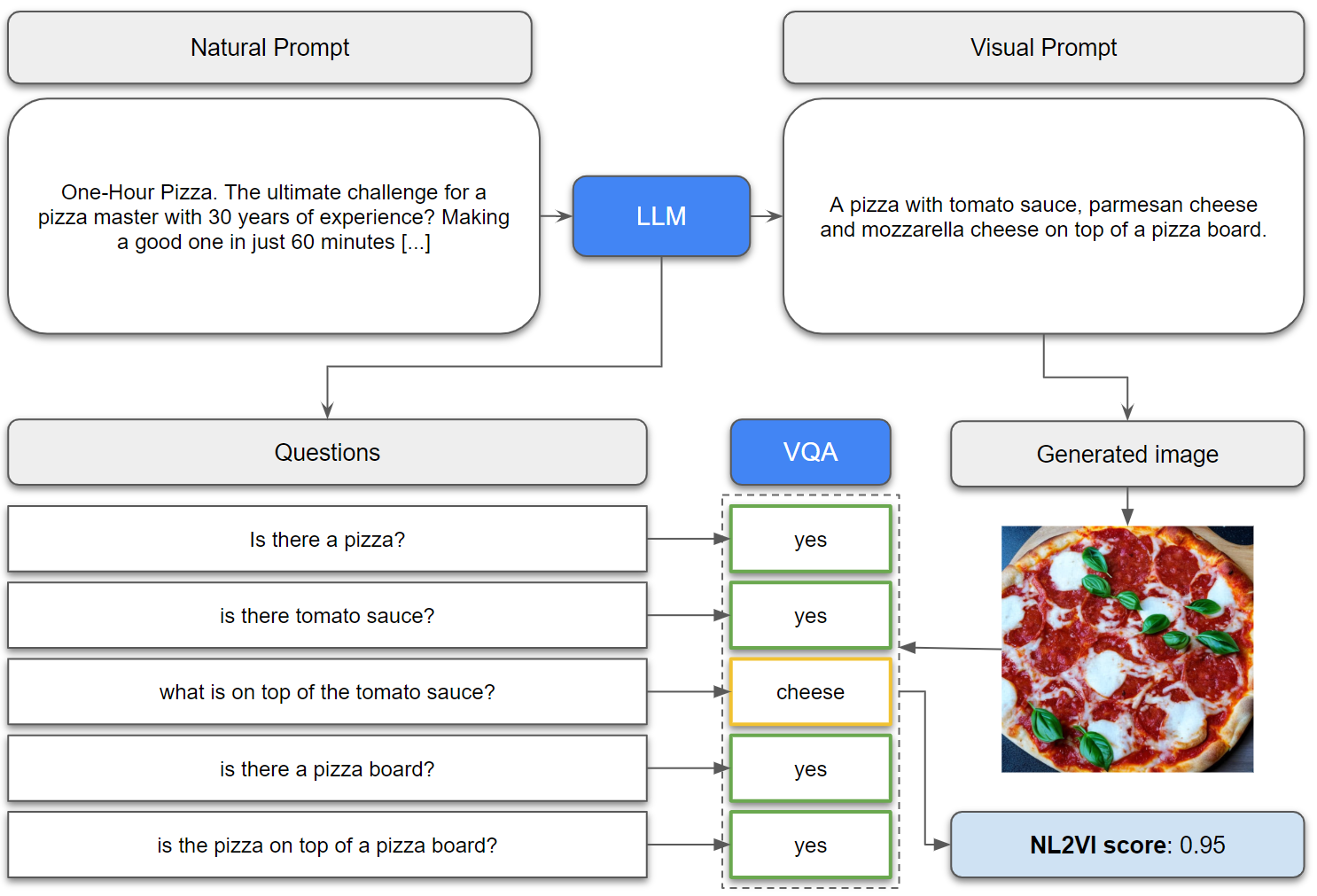}
    \caption{The NL2VI method implements an in-context few-shot learning approach: an LLM is responsible for generating the visual prompt and the verification questions for the QA and VQA methods.}
    \label{fig:NL2VI_example}
\end{figure}

The implementation of \approach (see Fig.~\ref{fig:NL2VI_example}) relies on many pre-trained models that are used during the different steps of the pipeline.
First, the natural prompt is transformed into a visual prompt and the consistency verification questions are generated, all through in-context learning. We experimented with two large language models: \texttt{gpt-3.5-turbo} from OpenAI and PaLM~540B~\cite{palm}. 
For reproducibility purposes, we release the visual prompts and generated questions, see Appendix~\ref{annex:dataset} for details.
Second, for the image generation step, we use Stable Diffusion 2.1, conditioned on the computed prompts, to generate the target image. 
While SD 2.1 is the current state-of-the-art, in the future, NL2VI can generalize to any other image generation method.
Third, the generated questions are answered by both QA and VQA methods based on the visual prompt and on the image together with the visual prompt, respectively. 
As \textbf{QA} filtering model, we tested the widely popular QANLU~\cite{qanlu} and the UnifiedQA model~\cite{unifiedqa} followed by the NLI model RoBERTa-NLI~\cite{nli-model}.
For \textbf{VQA}, we experimented with BLIP~\cite{blip}, GIT~\cite{git}, OFA~\cite{ofa}, PaLI~\cite{pali} and mPLUG~\cite{mplug}.

\subsection{Baselines}
\label{ssec:baselines}
NL2VI is a general approach that supports any T2I base method,  e.g. DALL-E2~\cite{dall-e2}, Imagen~\cite{imagen}. In the following experiments, we used Stable Diffusion 2.1, results with more methods can be found in the supplementary material. Moreover, the image verification step can be done with other methods, such as with CLIPScore~\cite{clipscore} or with the TIFA model~\cite{tifa}.

\subsection{NL2VI Public Dataset\footnote{Available after publication.}}
To study the generalization of T2I methods, we benchmark their performance in settings with natural prompts in the Recipes and WikiHow domains.
The \textbf{NL2VI natural prompts and questions} dataset comprises 3000 curated natural prompts, where the correct illustration of an action and correct composition of overlapping objects is both challenging and critical.
It was designed to allow for the realistic, yet controllable, research of image generation methods conditioned on natural prompts. 
See Annex~\ref{annex:dataset} for details.

\section{Results and Discussion}
Next, we present and discuss the NL2VI method's experimental results, highlighting the key takeaways.

\subsection{Verified Image Generation Results}
In this section, we compare the performance of NL2VI against other verified image generation approaches. We consider several LLMs and VQA methods, as well as the CLIPScore~\cite{clipscore} metric and the recent TIFA model~\cite{tifa}. 

\input{tables/vfc_accuracy.tex}

Table~\ref{tab:vfc-accuracy-table} summarizes the overall experimental results.
We observed that CLIPScore has a very low variance across all generated images, with a mean value of $\sim$32\% . We also noticed that this metric is not correlated with the visual consistency.
This might be explained by two factors: first, CLIPScore is already used by some T2I methods as the metric to be optimized, and second, CLIPScore does not capture fine-grained information between the image and the prompt, as discussed by related works~\cite{when-visionlanguage-bow}.
TIFA performs better than CLIPScore, achieving good results with explicit prompts, but fails when presented with more challenging natural language prompts. 
\approach is clearly superior to the other methods, specially with the combination of GPT-3.5 and PaLI. 
\approach was able to solve common inconsistencies present in other methods, w.r.t the lack of visual common sense knowledge.

\subsection{Visual Prompt Consistency}
\begin{figure}[t]
    \centering
    \begin{subfigure}[b]{0.49\linewidth}
    \includegraphics[width=\linewidth]{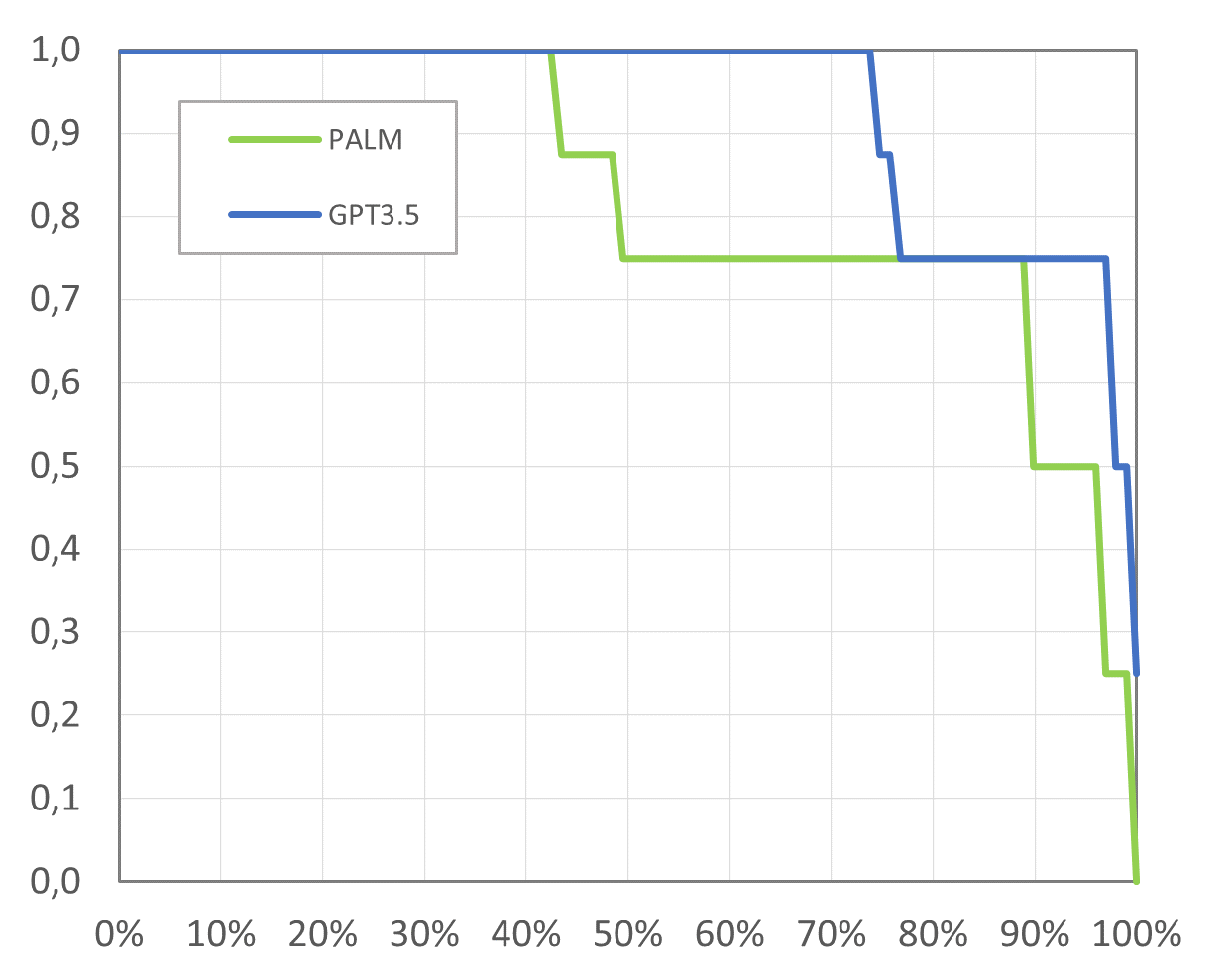}
    \caption{Recipes domain.}
    \label{fig:visual_prompt_consistency_recipes}
    \end{subfigure}
    \begin{subfigure}[b]{0.49\linewidth}
    \includegraphics[width=\linewidth]{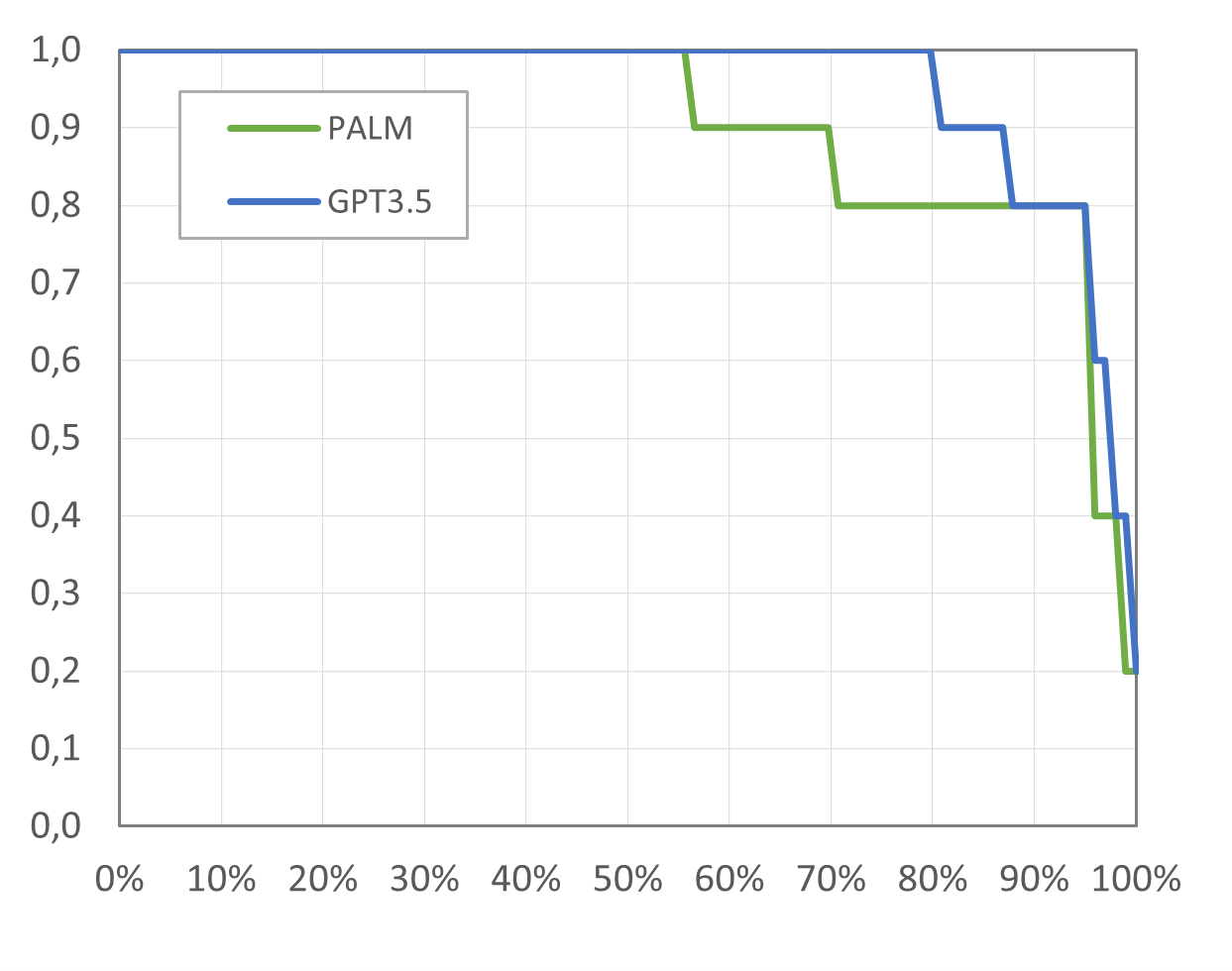}
    \caption{WikiHow domain.}
    \label{fig:visual_prompt_consistency_wikihow}
    \end{subfigure}
    \caption{Visual prompt consistency with respect to the natural prompt.}
    \label{fig:visual_prompt_consistency}
\end{figure}

In this Section, we analyse the consistency of the visual prompt with respect to its alignment with the natural prompt. The objective is to understand how well the visual elements of a natural prompt are unambiguously captured in the visual prompt. A human annotation task was set up for this purpose. Figure~\ref{fig:visual_prompt_consistency} presents the precision curve over the annotated corpus, and Table~\ref{tab:vfc-accuracy-table} presents summarized metrics of the curves.
In the Recipes domain, both LLMs perform exceedingly well, with GPT-3.5 achieving an average precision of 92.8\% and PaLM 82.5\%. 
This performance is even better in the WikiHow domain, with 90.2\% and 94.9\% average precision for PaLM and GPT-3.5, respectively.
In terms of precision at 1 (prompts that are fully correct), precision decreases in both models, with GPT-3.5 being able to generate visual prompts that could capture the visual elements of natural prompts in 74.0\% of the cases in the Recipes domain and 80.0\% of the cases in the WikiHow domain. 
This is a highly positive result, given that natural language prompts can depict a wide range of situations and concepts, which may not have an obvious visual representation. Appendix~\ref{annex:examples} illustrates some challenging examples.

\input{tables/visual_prompts.tex}

\subsection{Answer Verification Methods}
\begin{figure}[t]
    \centering
    \captionsetup[subfigure]{justification=centering}
    \begin{subfigure}[b]{0.48\linewidth}
        \centering
        \includegraphics[width=\linewidth]{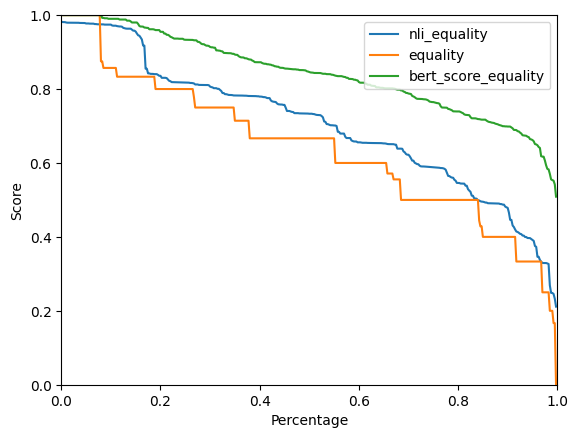}
        \caption{\tiny NL2VI (GPT-3.5).}
        \label{fig:a}
    \end{subfigure}
    \begin{subfigure}[b]{0.48\linewidth}
        \centering
        \includegraphics[width=\linewidth]{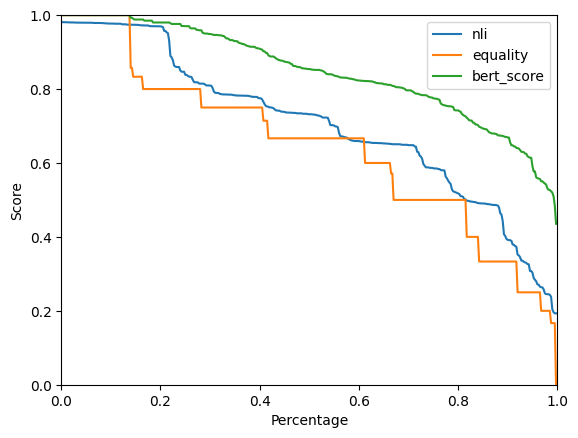}
        \caption{\tiny NL2VI (GPT-3.5).}
    \end{subfigure}
\\
    \begin{subfigure}[b]{0.48\linewidth}
        \centering
        \includegraphics[width=\linewidth]{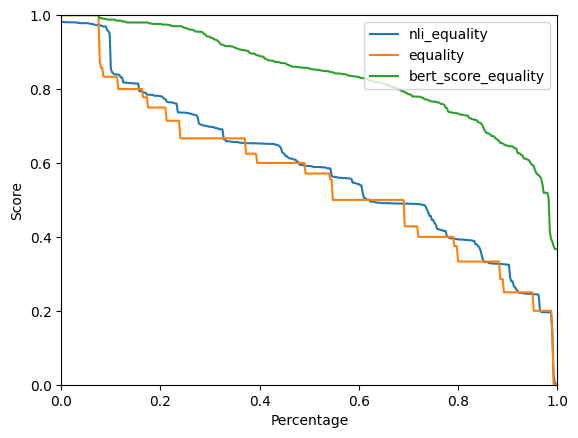}
        \caption{\tiny NL2VI (PaLM).}
        \label{fig:b}
    \end{subfigure}
    \begin{subfigure}[b]{0.48\linewidth}
        \centering
        \includegraphics[width=\linewidth]{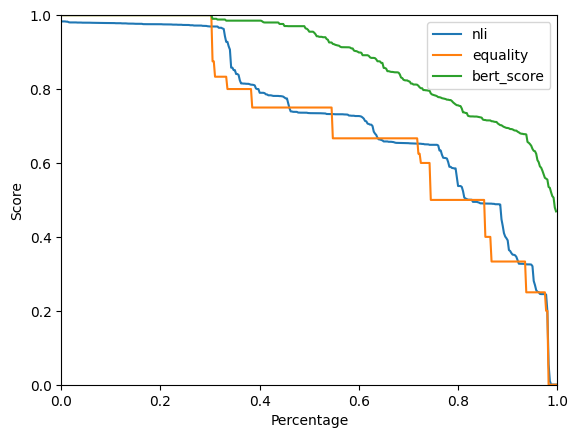}
        \tiny
        \caption{\tiny NL2VI (PaLM).}
    \end{subfigure}
\\    
    \begin{subfigure}[b]{0.49\linewidth}
        \centering
        \includegraphics[width=\linewidth]{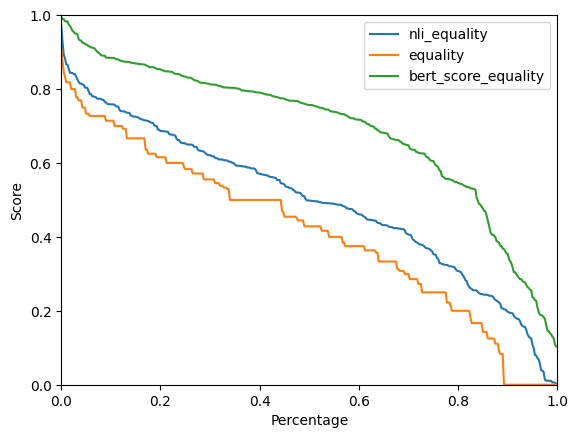}
        \caption{\tiny TIFA.}
        \label{fig:c}
    \end{subfigure}
    \begin{subfigure}[b]{0.49\linewidth}
        \centering
        \includegraphics[width=\linewidth]{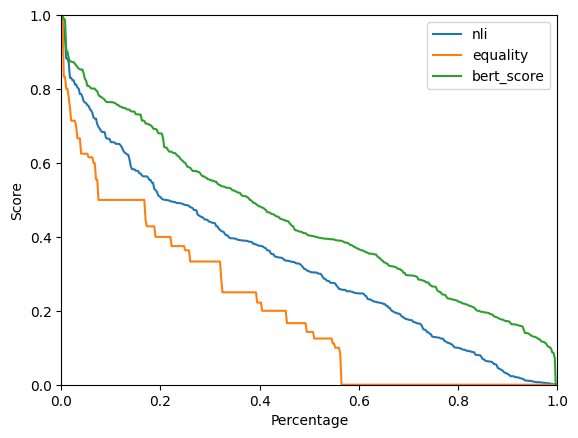}
        \tiny
        \caption{\tiny TIFA.}
    \end{subfigure}
    \caption{Image consistency on the Recipes (top row) and WikiHow (bottom row), according to the QA and VQA answers verification with string matching, NLI and BERTScore.}
    \label{fig:answer_verification}
\end{figure}

When comparing the VQA generative answers with the QA extractive answers, there are several discrepancies that need to be bridged due to the differences in the methods.
Computing the correspondence between answers with string matching algorithms would be rather limiting, which is why we investigated NLI~\cite{nli} and BERTScore~\cite{bertscore}, as alternatives.
Figure~\ref{fig:answer_verification} provides a visualization of the score distribution for each method. TIFA images which are based on natural prompts have, on average, lower scores than our images generated from visual prompts. Therefore, based on \approach, our images are more likely to be consistent with the prompt.

\begingroup
\begin{figure*}[t]
    \centering
    \begin{minipage}[t]{0.48\linewidth}
        \centering
        \begin{tabular}{@{}lccc@{}}
        \toprule
           & \textbf{Equality} & \textbf{NLI} & \textbf{BERT-Score} \\ \midrule
        \multicolumn{2}{l}{\textbf{TIFA}} \\\midrule
        \quad BLIP & 64.7 & 68.9 & 72.0 \\
        \quad GIT  & 62.4 & 67.3 & \textbf{72.5} \\
        \quad OFA  & 61.8 & 65.4 & \underline{72.2} \\
        \quad mPLUG & 64.3 & 67.4 & 71.1 \\
        \quad PaLI  & 61.9 & 66.5 & 68.3 \\ \midrule
        \multicolumn{2}{c}{\textbf{NL2VI w/ GPT-3.5}} \\\midrule
        \quad BLIP & 73.7 & 76.2 & 79.8 \\
        \quad GIT & 72.1 & 74.7 & 79.2 \\
        \quad OFA & 70.3 & 73.0 & \underline{79.9} \\
        \quad mPLUG & 74.3 & 75.9 & \underline{79.9} \\
        \quad PaLI  & 75.5 & 78.3 & \textbf{80.3}          \\ \midrule
        \multicolumn{2}{c}{\textbf{NL2VI w/ PaLM}} \\\midrule
        \quad  BLIP & 76.5 & 77.3 & \underline{78.2} \\
        \quad  GIT  & 72.8 & 74.1 & \textbf{78.4} \\
        \quad  OFA  & 70.1 & 70.9 & \textbf{78.4} \\
        \quad  mPLUG & 75.4 & 76.3 & 77.9 \\
        \quad  PaLI  & 75.7 & 76.5 & 77.7 \\ 
        \bottomrule
        \end{tabular}
        \captionof{table}{Ablation study on the Recipes domain: analysis of VQA and answer validation methods.}
        \label{tab:VQA-ablation-recipes}
    \end{minipage}
    \hfill
    \begin{minipage}[t]{0.48\linewidth}
        \centering
        \begin{tabular}{@{}lccc@{}}
        \toprule
            & \textbf{Equality} & \textbf{NLI} & \textbf{BERT-Score} \\ \midrule
        \multicolumn{2}{l}{\textbf{TIFA}} \\\midrule
        \quad BLIP & 56.1 & 61.6 & 64.2 \\
        \quad GIT & 56.1 & 62.7 & 64.8 \\
        \quad OFA & 56.3 & 62.7 & 64.2 \\
        \quad mPLUG & 58.4 & 64.3 & \underline{64.9} \\
        \quad PaLI  & 58.3 & \textbf{65.1} & 64.8 \\  \midrule
        \multicolumn{2}{l}{\textbf{NL2VI w/ GPT-3.5}} \\\midrule
        \quad          BLIP & 73.1 & 75.8 & 73.8 \\
        \quad         GIT & 71.0 & 73.9 & 73.9 \\
        \quad         OFA  & 71.6 & 73.6 & 73.6 \\
        \quad         mPLUG & 73.5 & \textbf{76.1} & 73.8 \\
        \quad         PaLI  & 74.0 & \underline{76.0} & 74.1 \\
        \midrule
        \multicolumn{4}{l}{\textbf{NL2VI w/ PaLM}} \\\midrule
        \quad   BLIP & 73.0 & 73.5 & 69.8 \\
        \quad         GIT & 70.8 & 71.3 & 69.7 \\
        \quad         OFA & 72.5 & \textbf{73.6} & 69.8 \\
        \quad         mPLUG & 72.3 & \underline{72.7} & 69.3 \\
        \quad         PaLI  & \underline{72.7} & 72.6 & 70.0 \\
        \bottomrule
        \end{tabular}
        \captionof{table}{Ablation study on the WikiHow domain: analysis of VQA and answer validation methods.}
        \label{tab:VQA-ablation-wikihow}
        \end{minipage}
\end{figure*}
\endgroup

Table~\ref{tab:VQA-ablation-recipes} and Table~\ref{tab:VQA-ablation-wikihow} illustrate the comparison between answer matching algorithms, as judged by human annotators.
The equality strategy is used for simple answers, like \textit{yes} and \textit{no}. NLI is used for longer answers, where entailment properties need to be checked.
BERTScore achieved the best performance on the Recipes domain, while NLI was the best in the WikiHow domain.
The first fact that stands out is that, independently of the method we use to validate the images, we observed that visual prompts (GPT-3.5 and PaLM) generate images that are better aligned with the natural prompt than images generated with the original natural prompt.
This is due to the linguistic ambiguities that may exist in the natural prompt and need to be solved by T2I algorithms. 
With visual prompts, these linguistic ambiguities were solved by an LLM, which is more capable of handling linguistic idiosyncrasies.
Furthermore, the visual prompt is more concise than the natural prompt and is therefore unaffected by the token limit of some T2I algorithms.
\input{tables/question_distribution.tex}

\input{tables/question_filtering.tex}

\begin{figure*}[t]
    \centering
    \includegraphics[width=0.8\textwidth]{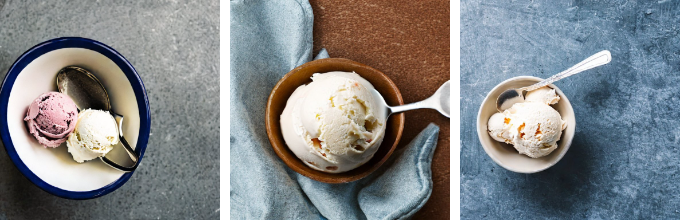}
    \caption{Images generated for "A bowl of ice-cream with a spoon". There is no information about the colour or type of ice-cream, and the model must fill in the missing properties. The same can be observed for the bowl itself. Another aspect is quantity, with the image on the left having two balls of ice-cream, which is information not present in the prompt, and the other images showing a single ball. We can also see how the spoons appear deformed.}
    \label{fig:open-world-example}
\end{figure*}

\subsection{Impact of QA and VQA performance}
\label{sec:ablation_qa_vqa}
In this Section, we report the results of an ablation study concerning the importance of the QA and VQA methods in the image consistency verification process. 
Table~\ref{tab:question-distribution} presents the statistics of the image verification questions that need to be answered by the QA and the VQA methods.
The key fact to note in this table is that the manual annotations of the generated questions show that 89.6\% and 87.4\% of the questions in the Recipes and WikiHow domains, respectively, are valid, thus confirming the overall quality of the generated questions.
Moreover, we can see that many questions are open-ended, which forces the use of generative VQA methods.
Table~\ref{tab:question-filtering-table} presents the results of two QA methods. 
QA methods were restricted to span-extraction methods and multiple-choice approaches, hence, the reason for using the QANLU~\cite{qanlu} and the UnifiedQA~\cite{unifiedqa} algorithms.
UnifiedQA was superior both in terms of precision and recall, and was the model chosen for our experiments.

With respect to VQA methods, we ran an extensive ablation study as presented on Table~\ref{tab:VQA-ablation-recipes} for the Recipes domain and Table~\ref{tab:VQA-ablation-wikihow} for the WikiHow domain. 
We cross-examined five VQA generative algorithms with natural prompts (TIFA) and visual prompts (PaLM and GPT-3.5) and three answer comparison methods: equality, NLI and BERTScore.
From these results, we can observe that mPLUG was always the best, or the second best performing method, in all settings. PaLI was the best, or second best, in four of the six experimental settings.
A key insight that we get from these results is that the performance of the VQA method is \textit{directly connected} to the performance of the visual consistency verification of the generated image.

\subsection{Hallucinations, Open-World Assumption and Visual Common Sense}
\label{ssec:hallucinations-closed-world-assumption-visual-common-sense}
Prompts fed to image generation models contain \textit{limited information}. It is not reasonable to assume that all the information that will be present in the final image was originally present in the prompt. Some common missing aspects in prompts are the image background and object colours and texture. Examples can be seen in appendix in Figure~\ref{fig:open-world-example}.
The lack of visual descriptions in the prompts, is filled in by model hallucinations, resulting in an image with more information than the original prompt. 
Hence, \textbf{hallucinations are needed} and unavoidable in the context of image generation, creating an \textbf{open-world setting} where multiple valid images can be generated from the original prompt.
Verifying image consistency in an open-world setting is a challenging task. 
In the present work, instead of verifying the visual consistency in an open-world setting, we chose to verify consistency based on what is present in the visual prompt, thus adopting a conditional closed-world assumption during verification.
LLMs are responsible to transform the open-world setting into a closed-world setting.
A key concern is guaranteeing that these hallucinations are aligned with \textbf{visual common sense}, as some hallucinations are plausible while others invalidate the consistency of an image. 
In the Recipes domain, most generated images correctly depict the prompt, even when they are complex, such as in Figure~\ref{fig:open-world-example} in the Appendix.
However, T2I algorithms sometimes lack common sense~\cite{t2i-polysemous}, especially when the meaning of the words is not clear, as in Figure~\ref{fig:open-world-example}. 
This issue falls out-of-scope of the present work, as we focus on verifying whether the elements present in the prompt show up in the final image.

\section{Related work}

Assessing the consistency of language or image generation methods is still an unsolved problem, despite having been addressed in the NLP field under different formulations, i.e., entailment~\cite{factual.consistency.entailment.summarization, factual.consistency.entailment.summarization.Falke}, counterfactual information~\cite{inconsistent-paraphrasing}, and question-answer approaches~\cite{q2,dialfact}.
Early work was done in the related tasks of face image hallucination~\cite{survey-face-hallucination} and image forensics to detect deepfakes and image tampering~\cite{survey-image-forensics}.
However, in the image generation domain, only after the publication of DALL-E~\cite{dall-e} and Stable Diffusion~\cite{stable-diffusion} models, has the community started to take the first steps towards assessing the visual consistency of T2I algorithms~\cite{seeing.double, dall-e2.fails.synthatic, t2i-polysemous, gokhale2022benchmarking, petsiuk2022human, park2021benchmark, russo2022creative}. 
The hallucinations and errors of T2I methods were recently discussed by \citet{seeing.double, dall-e2.fails.synthatic, t2i-polysemous, petsiuk2022human}, shedding some light on the lack of visual consistency between the generated image and the prompt.
An inspiring step is taken by~\citet{russo2022creative}, discussing possible methods to evaluate the artistic value of image generation methods. However, better methods for evaluating visual consistency are still lacking.
For example, traditional image quality metrics are not fine-grained enough, e.g., the inception score~\cite{inception_score} and FID~\cite{fid} are intended to measure the realism of the generated images and fail to catch inconsistencies~\cite{park2021benchmark}.
End-to-end evaluation of the similarity of image-text embeddings obtained with a dual-encoder multimodal model like CLIP~\cite{clip} fails to encode compositional information, which is crucial for visual consistency evaluation~\cite{embeddings_bag_of_words}. There have been attempts to improve such models, like CLIP-R~\citet{park2021benchmark} and CLIPScore~\cite{clipscore}, but results are still suboptimal.

Another line of work proposed several synthetic benchmarks and metrics to measure the visual consistency properties of T2I methods.
Early approaches relied upon object detection models and heuristics. \citet{gokhale2022benchmarking} proposed the VISOR metric, which evaluates the spatial relationship between objects detected in the scene. Similarly, \citet{dalleval} proposed the PaintSkills dataset to assess object presence, properties and the spatial relationships described in the prompt.
Drawbench is another benchmark introduced by Imagen~\cite{imagen}.
A more generic approach was proposed by~\citet{tifa}, which follows the literature of natural language factual consistency with QA~\cite{true, feqa}. In this method, several questions are generated from the prompt and verified in the image with a VQA method.
Concurrent to our work, ~\citet{yarom2023read} introduces the SeeTRUE benchmark for meta-evaluation of image-text alignment.
While all these methods can measure T2I average correctness with synthetic or visually descriptive prompts, they are not designed to correctly generate images conditioned on natural prompts.

\section{Conclusions}
\paragraph{Contributions.}
Generating images from natural language that is non-visual is, in many cases, an impossible task for image generation methods, like Stable Diffusion. In this context, the key contributions of this paper are twofold.

First, the \approach method to transfer the visual attributes of a natural language into a visual prompt that will generate a verified image. 
The visual prompt is correctly aligned with the natural prompt in over 90\% of the cases and is also an enabler of a verification process based on VQA methods.
The image verification step is the final safeguard to ensure that the image was correctly generated, according to the initial text.

Second, a \textit{public dataset} with natural prompts, visual prompts and verification questions to benchmark image generation methods in the presence of natural language. The curated dataset aggregates a series of natural language prompts that are from the instructions domain with descriptions that are not always visual.

\paragraph{Limitations.}
As discussed in Section~\ref{ssec:hallucinations-closed-world-assumption-visual-common-sense}, this work assumes a closed-world setting when verifying the consistency of an image. Although our method improves consistency, image generation works in an open-world setting, and by following a prompt-based verification, we will be missing some important factors of consistency. To this end, the verification is limited by the information present in the prompt, which limits verification. Furthermore, when generating visual prompts, some elements may be added, which were not present, originally. This can lead to some differences between the expected result and \approach's output. There is still important work to be done to be able to support a more broad verification of generated images.

\paragraph{Broader Impacts.} By improving the consistency of an image generation process, we are trying to more faithfully depict what is described in the original prompt.
The most obvious adverse impact, is the malicious use of generative image generation for disinformation or deceiving. 
We are against the applications of generative AI non-ethical uses and argue for a responsible and accountable use of these algorithms.

\section*{Acknowledgments} This work was partially funded by a Google Cloud gift and the NOVA LINCS research center, Ref. UIDP/04516/2020.

\bibliography{main}
\vspace{40mm}

\appendix

\vspace{5mm}
\section{NL2VI Public Dataset}
\label{annex:dataset}
To study the generalization of T2I methods, we benchmark their performance in settings with natural prompts in the Recipes and WikiHow domains.
This dataset was designed to allow for the realistic, yet controllable, research of image generation methods conditioned in natural prompts.

\subsection{Statistics}
The \textbf{NL2VI natural prompts and questions} dataset comprises 3000 curated natural prompts from the instructions domain, where the correct illustration of an action and correct composition of overlapping objects is critical. 

\subsection{Manual Annotation}
\label{annex:annotation_tool}
\begin{figure*}[t!]
    \centering
    \includegraphics[width=0.9\textwidth]{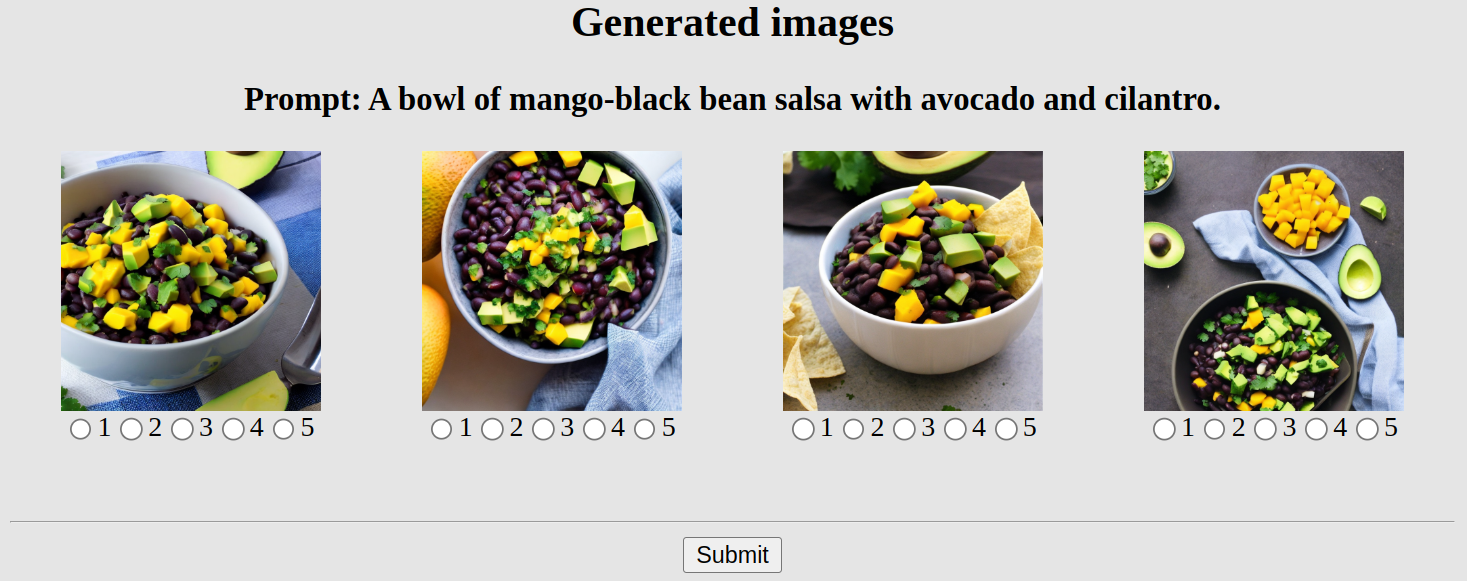}
    \caption{Image consistency annotation tool.}
    \label{fig:annotation-tool-recipe-images}
\end{figure*}

In order to collect human data to use as a baseline, we created an annotation tool. As seen in Figure~\ref{fig:annotation-tool-recipe-images}, labellers are shown 4 images, generated for a single prompt, and asked to rate them from 1 to 5. A score of 1 means the image was \textit{Not Consistent} with the prompt, a score of 3 means it was \textit{Somewhat Consistent} with the prompt, and a score of 3 means it was fully \textit{Consistent} with the prompt. 
These annotations were collected for the all the methods we tried.

\subsection{Verified Images Examples}
\label{annex:examples}
\begin{figure*}[th]
    \centering
    \includegraphics[width=0.7\textwidth]{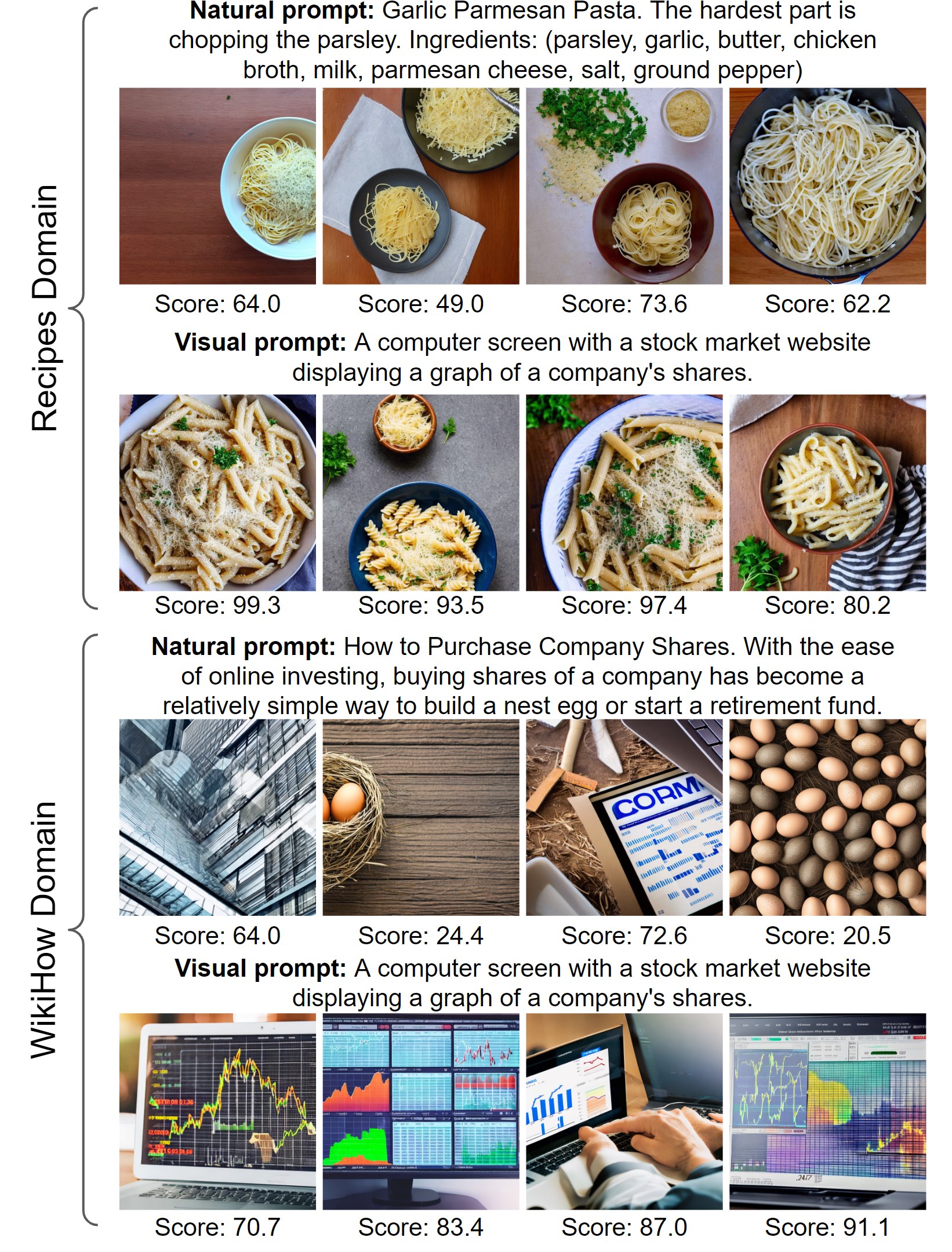}
    \caption{Comparison of natural and visual prompts for image generation across different domains.}
    \label{fig:open-world-example2}
\end{figure*}

Figure~\ref{fig:open-world-example2} presents examples of the \approach method.
On the Recipes domain, the images from the natural prompt miss important visual ingredients, like parsley. The visual prompt improves the consistency and consists only of relevant visual information.
On WikiHow, the natural prompt is penalized by being larger than the maximum sequence length allowed by Stable Diffusion. Moreover, it consists mostly of non-visual information, which degrades image generation. The visual prompt is able to overcome these problems.
\clearpage

\end{document}

%% file: tables/vfc_accuracy.tex
\begin{table}[t]
    \small
    \begin{tabular}{@{}lll|c|c@{}}
        \toprule
        \textbf{Models} & \textbf{LLM} & \textbf{VQA} & \textbf{Recipes} & \textbf{Wikihow} \\\midrule
        CLIPScore  & n/a     & n/a     & 57.4                        & 53.8                       \\
        TIFA     & GPT-3.5  & mPLUG   & 72.5                        & 64.9                 \\
        NL2VI      & PaLM    & OFA    & 78.4                        & 73.6                  \\
        NL2VI      & GPT-3.5  & PaLI    & 80.3                        & 76.0                  \\
        \bottomrule
    \end{tabular}
    \centering
    \caption{\label{tab:vfc-accuracy-table} Human evaluation of the different visual factual consistency methods.}
\end{table}

%% file: tables/visual_prompts.tex
\begin{table}[t]
        \centering
        \begin{tabular}{@{}l|cccc@{}}
            \toprule
            \multirow{2}{*}{\textbf{Models}}  & \multicolumn{2}{c}{\textbf{Recipes}} & \multicolumn{2}{c}{\textbf{Wikihow}} \\
                           & AUC   &   P@1      & AUC     &  P@1   \\  \midrule
            PaLM           & 82.5  &   42.0     & 90.2    &  56.0  \\
            GPT-3.5        & 92.8  &   74.0     & 94.9    &  80.0  \\
            \bottomrule
        \end{tabular}
        \captionof{table}{Alignment between natural language prompts and visual generation prompts. }
        \label{visual-prompts-table}
\end{table}

%% file: tables/question_distribution.tex

\begin{table}[h]
\centering
\begin{tabular}{lcc}
\toprule
 &
  \textbf{\begin{tabular}[c]{@{}c@{}}Generated\\ Questions\end{tabular}} &
  \textbf{\begin{tabular}[c]{@{}c@{}}Verified\\ Questions\end{tabular}} \\ \midrule
\textbf{Recipes}\\
\midrule
\quad Binary    & 392 & 386 \\
\quad Open-Ended & 233 & 174 \\
\midrule
\textbf{Wikihow}\\
\midrule
\quad Binary      & 403 & 388 \\
\quad Open-Ended  & 222 & 158 \\ 
\bottomrule
\end{tabular}
\caption{Distribution of generated questions for the Recipes and WikiHow datasets before and after filtering (UnifiedQA).}
\label{tab:question-distribution}
\end{table}

%% file: tables/question_filtering.tex
\begin{table*}[h]
\centering
\begin{tabular}{@{}l|cc|cc|cc@{}}
\toprule
 & \multicolumn{2}{c|}{\textbf{\% of valid questions}} & \multicolumn{2}{c|}{\textbf{Precision}} & \multicolumn{2}{c}{\textbf{Recall}} \\ \midrule
\textbf{Question filtering}  & \textbf{Recipes}         & \textbf{Wikihow}        & \textbf{Recipes}       & \textbf{Wikihow}     & \textbf{Recipes}     & \textbf{Wikihow}   \\ \midrule
QANLU                       & 50.4                     & 50.8                    & 22.2                   & 33.3                  & 28.6                 & 28.6               \\
Unified-QA                  & 89.6                     & 87.8                    & 90.9                   & 84.2                  & 71.4                 & 76.2               \\ \bottomrule
\end{tabular}
\caption{\label{tab:question-filtering-table}Evaluation of the question filtering stage.}
\end{table*}
